\documentclass[lettersize,journal]{IEEEtran}
\usepackage[ruled, vlined, linesnumbered]{algorithm2e}
\usepackage{amsmath,amsfonts}
\usepackage{xcolor,soul,framed}
\usepackage{algorithmic}
\usepackage{array}
\usepackage[caption=false,font=normalsize,labelfont=sf,textfont=sf]{subfig}
\usepackage{textcomp}
\usepackage{stfloats}
\usepackage{url}
\usepackage{verbatim}
\usepackage{multirow}
\usepackage{graphicx}
\usepackage{cite}
\usepackage{booktabs}
\def\red#1{\textcolor{black}{#1}}
\newenvironment{tred}{\par\color{black}}{\par}

\begin{document}

\title{PerFED-GAN: Personalized Federated Learning \\ via Generative Adversarial Networks}

\author{Xingjian Cao,
        Gang Sun,
        Hongfang Yu,
        Mohsen Guizani

\IEEEcompsocitemizethanks{
\IEEEcompsocthanksitem This research was partially supported by the National Key Research and Development Program of China (2019YFB1802800), PCL Future Greater-Bay Area Network Facilities for Large-scale Experiments and Applications (PCL2018KP001).
\IEEEcompsocthanksitem X. Cao is with the Key Laboratory of Optical Fiber Sensing and Communications, Ministry of Education, University of Electronic Science and Technology of China, Chengdu 6111731, China (e-mail: x.cao.flc@gmail.com).
\IEEEcompsocthanksitem G. Sun is with the Key Laboratory of Optical Fiber Sensing and Communications, Ministry of Education, University of Electronic Science and Technology of China, Chengdu 6111731, China, and is also with the Agile and Intelligent Computing Key Laboratory of Sichuan Province, Chengdu, 6111731, China (e-mail: gangsun@uestc.edu.cn).
\IEEEcompsocthanksitem H. Yu is with the Key Laboratory of Optical Fiber Sensing and Communications, Ministry of Education, University of Electronic Science and Technology of China, Chengdu 6111731, China, and is also with the Pengcheng Laboratory, Shenzhen 518000, China (e-mail: yuhf@uestc.edu.cn).
\IEEEcompsocthanksitem M. Guizani is with the Machine Learning Department, Mohamed Bin Zayed University Of Artificial Intelligence (MBZUAI), Abu Dhabi, UAE (e-mail: mguizani@ieee.org).}% <-this % stops a space
}

\maketitle

\begin{abstract}
\red{Federated learning is gaining popularity as a distributed machine learning method that can be used to deploy AI-dependent IoT applications while protecting client data privacy and security. Due to the differences of clients, a single global model may not perform well on all clients, so the personalized federated learning method, which trains a personalized model for each client that better suits its individual needs, becomes a research hotspot. Most personalized federated learning research, however, focuses on data heterogeneity while ignoring the need for model architecture heterogeneity. Most existing federated learning methods uniformly set the model architecture of all clients participating in federated learning, which is inconvenient for each client's individual model and local data distribution requirements, and also increases the risk of client model leakage. This paper proposes a federated learning method based on co-training and generative adversarial networks(GANs) that allows each client to design its own model to participate in federated learning training independently without sharing any model architecture or parameter information with other clients or a center. In our experiments, the proposed method outperforms the existing methods in mean test accuracy by 42\% when the client's model architecture and data distribution vary significantly.}
\end{abstract}


\section{Introduction}
\red{\IEEEPARstart{I}{nternet} of Things (IoT) devices are growing dramatically. It is estimated that in the next 10 years, the total number of IoT devices will reach hundreds of billions\cite{IHS-Markit}, and these devices are used in fields such as intelligent transportation\cite{liu2020privacy}, intelligent industry\cite{kang2020reliable}, healthcare and intelligent monitoring\cite{khan2020edge}, \cite{sanchez2019smart}. These IoT devices generate massive and rapidly growing data, and present attractive opportunities for data-driven artificial intelligence techniques\cite{qi2020popularity}. Centralized machine learning schemes can be used 
for these smart IoT applications. However, centralized machine learning techniques have an inherent problem of leaking user privacy because end-device data needs to be transferred to a centralized third-party server for training. Furthermore, centralized machine learning may not be feasible when the amount of data is very large and located in multiple locations. Although distributed machine learning can solve some problems of centralized machine learning, traditional distributed machine learning methods still lack the protection of data security and privacy of participants.}

\red{Federated learning is a distributed machine learning method proposed in \cite{mcmahan2017communication}, and it can train machine learning models using distributed data stored in different clients while avoiding client data leakage. The clients in the scenario where federated learning was first proposed are various edge devices such as mobile phones\cite{konevcny2016federated}, \cite{liu2020privacy}. The center trains a neural network global model using the data stored in each client device. The center creates the global model architecture, which is shared by all clients. The need for personalized federated learning was then proposed\cite{jiang2019improving}, \cite{arivazhagan2019federated}, \cite{kairouz2019advances}, \cite{liu2020federated}, where the needs and data distribution of individual clients are typically quite different. For example, each client needs to train a shopping recommendation model for its customers, and each client's customer groups are geographically diverse. A global model is difficult to fit all clients due to factors such as different consumer preferences and spending power in different regions, and the data distribution of different clients may also be non-independent and identically distributed (\red{Non-IID}). As a result, the goal of personalized federated learning is to train a personalized model for each individual client that is more tailored to their specific needs, rather than a single global model that serves all clients.}

Although many personalized federated learning methods have been proposed, the majority of these methods require all clients' model architectures to be consistent, and model personalization is reflected in differences in model parameters. \red{However, when the clients participating in federated learning belong to different organizations, the differences in client models are likely not limited to the parameters of the models, but also include different model architectures. Because the machine learning model architectures deployed by different organizations on their client devices are designed for their specific task requirements, the client model architectures of different organizations are typically not the same. We believe that the consistency of the model architecture limits the scope of personalized models. Additionally, sharing the model architecture results in the leakage of the model design, which may be the intellectual property rights of participating organizations. Therefore, if the model architecture is known, maliciously participating clients will be more involved. It is simple to use the model parameters and training process to infer other clients' private training data\cite{mothukuri2021survey}, \cite{ren2021grnn}, and this invalidates the role of federated learning in protecting client privacy.} 

The effectiveness of federated learning stems from knowledge sharing among clients, and personalized federated learning aims to share common knowledge among clients while maintaining their personalized knowledge. \red{Existing federated learning methods share knowledge via model parameters, and necessitate consistent model architectures, which not only weakens the personality of the client model but also increases the risk of model leakage.} This paper proposes a federated learning strategy based on generating adversarial sample sharing to overcome this issue. The proposed strategy allows each client to independently design the neural network model architecture without disclosing it to other clients or the center. The knowledge of each client's training data does not need to be shared in the form of model parameters, but can be obtained by sharing the samples generated by generated adversarial networks (GANs). Our main contributions are as follows:
\begin{itemize}
\item \red{We propose a personalized federated learning method where clients train personalized models for their individual needs.}
\item The proposed method is applicable to federated learning client neural network models of various architectures and outperforms existing methods for \red{Non-IID} data.
\item \red{We theoretically analyze the convergence of the proposed method and derive its convergence conditions.}
\item \red{We conduct many experiments to validate the efficacy of the proposed method and analysis conclusions.}
\end{itemize}

\red{The remainder of this paper is organized as follows. We briefly review the related studies in Section II. We formulate the problem statement in Section III. In Section IV, we describe the proposed PerFED-GAN method in details. Section V presents the experimental details and results, and Section VI summarizes this paper.}

\section{Related Work}
\red{\subsection{Personalized Federated Learning}}
In 2017, McMahan \textit{et al.} \cite{mcmahan2017communication} proposed the Federated averaging (FedAvg) algorithm to solve the federated optimization problem on edge devices. In the FedAvg method, a central server orchestrates the following training process. The central server broadcasts the global model to the selected clients (edge devices). \red{Then, the selected clients locally update their models by running stochastic gradient descent (SGD) on their local private datasets.} After that, The central server collects an aggregate of the client updates and updates the global model by averaging these aggregated updates. 

This process template for federated learning training that encompasses FedAvg and \red{many of its variants} by Wang \textit{et al.} \cite{wang2019adaptive} and Li \textit{et al.} \cite{li2018federated}, works well for many federated learning settings, where all clients usually serve a unified task and model designed by the center, and the center finely controls local training options (e.g. learning rate, number of epochs and mini-batch size). However, these methods are not compatible with personalized tasks, models, or training of different clients. For example, FedProx by Li \textit{et al.} \cite{li2018federated} introduces proximal terms to improve FedAvg in the face of system heterogeneity (i.e., many \red{stragglers}) and statistical heterogeneity, but it still inherits FedAvg polymerization parameters manner, therefore is not compatible with heterogeneous models, especially different architecture models.

In recent years, many efforts have been made to tackle the personalized tasks of clients in federated learning settings. Wang \textit{et al.} \cite{wang2019federated} and Jiang \textit{et al.} \cite{jiang2019improving} use the federated learning trained model as a pretrained or meta-learning model, and fine-tune the model to learn the local task of each client. Arivazhagan \textit{et al.} \cite{arivazhagan2019federated} use personalized layers to specialized the global model trained by federated learning to learn the personalized tasks of clients. Recently, Hanzely \textit{et al.} \cite{hanzely2020federated} and Deng \textit{et al.} \cite{deng2020adaptive} modify the original FedAvg, instead of aggressive averaging all the model parameters, they find that only steering the local parameters of all clients steps towards their average helps each client to train its personalized model. Besides, Smith \textit{et al.} \cite{smith2017federated} propose a federated multi-task learning framework MOCHA, which clusters tasks based on their relationship by an estimated matrix. Huang \textit{et al.} \cite{huang2021personalized} propose a personalized cross-silo federated learning method FedAMP by a novel attentive message passing mechanism that adaptively facilitates the underlying pairwise collaborations
between clients by iteratively encouraging similar clients to collaborate more.

However, all these methods require clients to upload their model parameters for global aggregation, which may leak {client models}. Recently, several secure computing techniques have been introduced to protect data and model parameters in federated learning, including differential privacy \cite{wei2020federated}, \cite{hu2020personalized}, \cite{xin2020private}, secure multi-party computing \cite{yin2020fdc}, \cite{liu2020secure}, \cite{zhu2020privacy}, homomorphic encryption \cite{gao2019privacy}, \cite{soe2020homomorphic}, and trusted execution environments \cite{mo2019efficient}, \cite{chen2020training}, but these methods still have some disadvantages, such as a significant amount of communication or computational cost, \red{or relying on specific hardware for implementation.} 

Another restriction of these federated learning approaches for personalized tasks is that the architecture of each client's model must be consistent because model parameters are typically aggregated or aligned during the federated learning training processes of these methods. It prevents clients from independently designing unique model architectures. To address this issue, FedMD \cite{li2019fedmd} proposed by Li \textit{et al.}, leverages knowledge distillation to convey the knowledge of each client's local data, by aligning the logits of multiple neural networks on a public dataset. FedMD's client models can be neural networks of various architectures, as long as they are consistent in the logits output layers. However, FedMD requires a large amount of labeled data as a public dataset for client models alignment. Collecting labeled data is difficult, and open large-scale labeled datasets are usually rare, which limits the application scenarios of FedMD. Similarly, Guha \textit{et al.} \cite{guha2019one} use distillation models to generate a global ensemble model, which needs to share the clients' local models or their distillation models. However, it exposes the clients' private data or local model information to the central server, \red{which makes this solution unsuitable for customers who want to avoid model leakage.}

\red{\subsection{Co-training}}
\red{Co-training, which is originally proposed by Blum \textit{et al.} \cite{blum1998combining}, is a semi-supervised learning technique exploiting unlabeled data with multi-views. The \textit{view} refers to a subset of attributes of an instance. It trains two different classifiers on two views; after that, both classifiers pseudo-label some unlabeled instances for another; then each classifier can be retrained on its original training set and the new set pseudo-labeled by another. Co-training repeats the above steps to improve the performance of classifiers until it stops. Blum \textit{et al.} proves that when the two views are sufficient (i.e., each view contains enough information for an optimal classifier), redundant and conditionally independent, co-training can improve weak classifiers to arbitrarily high by exploiting unlabeled data. Wang \textit{et al.} \cite{wang2013co} prove that if the two views have large diversity, co-training suffers little from the \textit{label noise} and \textit{sampling bias}, and could output the approximation of the optimal classifier by exploiting unlabeled data even with insufficient views.}

\red{For the single view setting, Zhou \textit{et al.} \cite{zhou2005tri} propose the tri-training, which uses 3 classifiers having large diversity to vote for the most confident pseudo-label of unlabeled data. Furthermore, Wang \textit{et al.} \cite{wang2007analyzing} prove that classifiers with large diversity can also improve model performance in a single view setting. At the same time, they also point out the reason why the performance of classifiers in co-training saturates after some rounds. That is, classifiers learn from each other and gradually become similar, and their differences are insufficient to support further performance improvement.}

\red{Different clients in personalized federated learning settings typically have insufficient local data that is solely one-sided for the total distribution, especially in \red{Non-IID} data settings. As a result, models trained on local data from multiple clients may have a high degree of variety, which can lead co-training to work.}

\red{\subsection{Generative Adversarial Network}}
\red{Generative adversarial network (GAN) \cite{goodfellow2014generative} is a type of machine learning framework designed by Ian Goodfellow and his colleagues in 2014. For a given training set, the technique learns to generate new data samples with the same distribution of it.}

\red{GAN is mainly composed of two parts: a discriminator network and a generator network. The core idea of GAN is to train indirectly through the discriminator network, and the discriminator network itself is also dynamically updated. This basically means that the generator network is not trained to minimize the distance to a particular object, but is used to fool the discriminator network.}

\red{The generating network generates candidates, and the discriminator network evaluates them. The competition is conducted in terms of data distribution. Generally, the generative network learns to map from the latent space to the data distribution of interest, and the discriminator network distinguishes the candidates generated by the generator from the real data distribution. The training goal of the generator network is to increase the error rate of the discriminator network.}

\red{The initial training dataset is used for the discriminator network. The training process involves showing it the samples from the training dataset. The generator is trained according to whether it succeeds in fooling the discriminator. Usually, the seed of the generator is a random input sampled from a predefined latent space. Thereafter, the candidates synthesized by the generator are evaluated by the discriminator.}

\red{\section{Problem Statement}}
\red{Considering a personalized federated learning setting, there are $N$ clients, and each client $C_i$($1 <= i <= N$) has its private dataset $D_i$ from distributions $\mathcal{D}_i$. Since different client data in personalized federated learning scenarios are generally \red{Non-IID}, that is:
\begin{equation}
\label{p1}
\exists i \ne j (1 <= i, j <= N) \rightarrow  \mathcal{D}_i\ne \mathcal{D}_j
\end{equation}
Each $C_i$ also has its model $M_i(w_i)$ where $M_i$ is the model architectures and $w_i$ is the corresponding model parameters of $M_i$. In a personalized federated learning scenario compatible with heterogeneous model architectures, we assume that the model architectures of different clients can be different, that is:
\begin{equation}
\label{p2}
\exists i \ne j (1 <= i, j <= N) \rightarrow  M_i\ne M_j
\end{equation}
At this time, the $w_i$ of $M_i$ and the $w_j$ of $M_j$ are also different. The difference is not only the difference in parameter values, but may even be the number of parameters, because different model architectures are likely to have different number of model parameters. Even in the coincidental case where the parameter number of $M_i$ and $M_j$ are equal, the parameter values of $w_i$ and $w_j$ have different meanings, so comparing their values makes no sense.}

\red{Define a function $f(x;M_i,w_i)$ to evaluate the performance of the model $M_i(w_i)$ for the sample $x$, the smaller the value of the $f$ function implies the better the model performance for $x$. For a classification task, when $x$ is correctly classified by $M_i(w_i)$, the output of $f$ is 0, otherwise it is 1. For a client $C_i$, its task objective is to optimize $w_i$ for its given model architecture $M_i$ so as to minimize the expectation of the function $f$ on $\mathcal{D}_i$:
\begin{equation}
    \label{p3}
    w_i^{*} = \underset{w_i}{arg\min}  \underset{x\sim\mathcal{D}_i}{\mathbf{E}}\left[ f\left( x;M_i,w_i \right) \right] 
\end{equation}
The $\mathbf{E}(\cdot)$ is the expectation function. For classification tasks, Formula (\ref{p3}) means optimizing $w_i$ to minimize the generalization classification error of the model $M_i$. Assume that the optimal model parameters obtained by local training of client $C_i$ are $w_{i,loc}^{*}$ for $i = 1, 2, \cdots, N$:
\begin{equation}
    \label{p4}
    w_{i,loc}^{*} \overset{local}{=} \underset{w_i}{arg\min}  \underset{x\sim\mathcal{D}_i}{\mathbf{E}}\left[ f\left( x;M_i,w_i \right) \right]
\end{equation}
The aim of personalized federated learning is to training a personalized model $M_i(w_i)$ for each client collaboratively, without exposing $D_i$ to $C_j$($1<=i, j<=N$, $i \ne j $):
\begin{equation}
    \label{p5}
    w_{i,fed}^{*} \overset{federated}{=} \underset{w_1,w_2,\cdots ,w_N}{arg\min}\,\,\frac{1}{N}\sum_{i=1}^N{\underset{x\sim \mathcal{D} _i}{\mathbf{E}}\left[ f\left( x;M_i,w_i \right) \right]}
\end{equation}
The performance of the model obtained by federated training should be sufficiently high. Specifically, the model performance of each client in federated learning should not be lower than the local training model performance for $i = 1, 2, \cdots, N$:
\begin{equation}
    \label{p6}
    \underset{x\sim\mathcal{D}_i}{\mathbf{E}}\left[ f\left( x;M_i,w_{i,fed}^{*} \right) \right] <= \underset{x\sim\mathcal{D}_i}{\mathbf{E}}\left[ f\left( x;M_i,w_{i,loc}^{*} \right) \right]
\end{equation}
For classification tasks, that is, the classification generalization accuracy of the federated training model of each participant is not lower than that of the local training model.}

\hfill
\section{PerFED-GAN}\label{sec4}
In this part, we introduce the main ideas of the PerFED-GAN method for personalized federated learning training, and analyze its algorithm complexity and convergence.

\subsection{The Overall Steps of PerFED-GAN}
The primary motivation for federated learning is to increase client model performance. The reason for the poor performance of the locally trained model is frequently a lack of local data, which prevents the model from learning enough task expertise by training on only the local dataset. As a result, in order to increase the performance of the client models, they must be able to learn from other clients. The most popular and direct ways to communicate knowledge are to share data or models, however these are disallowed in our federated learning environment, \red{therefore we need to develop alternative methods.}

\begin{figure}
\centering
\includegraphics[width=3.4in]{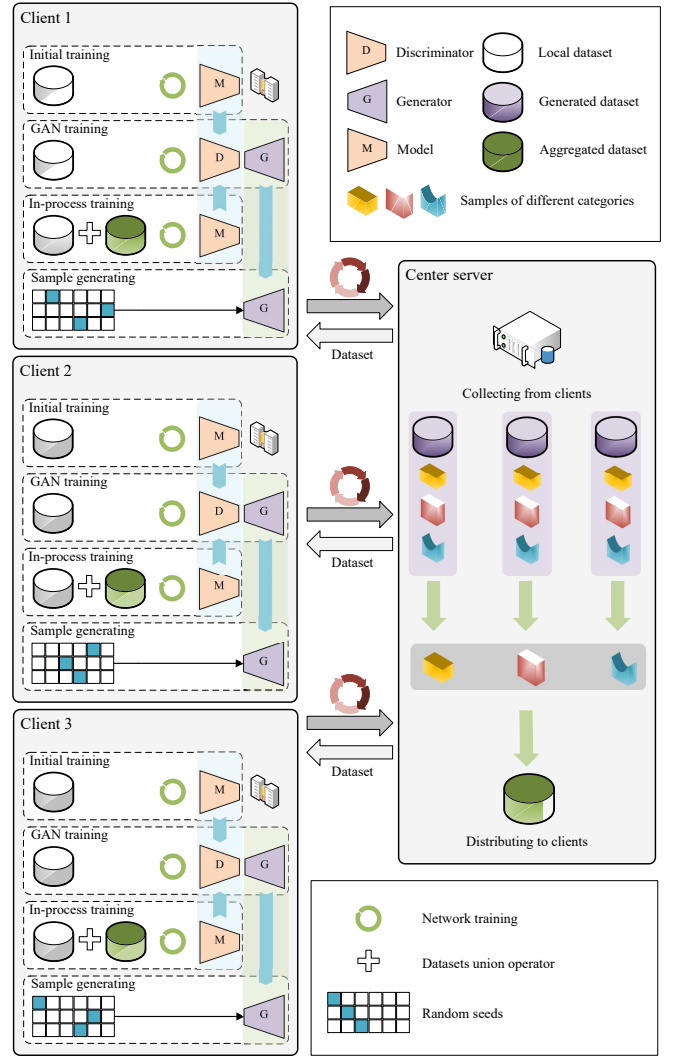}
\caption{Framework of PerFED-GAN (3 clients example).}
\label{lct}
\end{figure}

The study related co-training shows that, to improve the performance of the classification task model, a large enough diversity between models is required \cite{wang2007analyzing}. In general, generating models with substantial divergences in single view settings is difficult. In federated learning contexts, however, client models may have quite distinct architectures and be trained on individualized private datasets that are very likely to be \red{Non-IID}. All of these factors may result in variations in the models of different clients. As a result, our central idea is to use the model trained on each client's local data as the discriminator network, and to generate fresh samples. Following the receipt of these created datasets, other clients use them to train their local models further, thereby boosting the performance of their tailored models. We give the overall steps of PerFED-GAN as follows and Fig.\ref{lct}.

\begin{enumerate}
    \item \textbf{Local model training}: Each client trains independently its model on its local dataset.
    \item \textbf{GAN training}: Each client uses the local model trained in the previous step as a discriminator network to train a generator network, and uses it to generate a new dataset.
    \item \textbf{Communication}: After that, the center aggregates the generated data samples collected from the clients according to the label, and sends the results back to each client.
    \item \textbf{\red{Client model updating}}: Each client updates its model by training it on the new dataset merged from its local dataset and the received new dataset. After that, Step 2) to Step 4) are executed in a loop to further improve the performance of the personalized federated learning model.
    
\end{enumerate}

\begin{tred}
It should be pointed out that the local model training method and the GAN training method used in PerFED-GAN are modular and replaceable. In theory, almost all neural network training methods and GAN training methods can be applied to PerFED-GAN.

After the GAN training stage, the central server collects all categories of data from the data generated by each client during the aggregation process. For example, there are 3 categories, A, B, and C. Client 1 generates a dataset \{A1, A2, B1, B2, C1, C2\}, Client 2 generates a dataset \{A3, A4, B3, B4, C3, C4\}, and Client 3 generates a dataset \{A5, A6, B5, B6, C5, C6\}. The central server aggregates these to a large dataset \{A1, A2, …, A6, B1, B2,…,B6, C1, C2,…,C6\}, and then, the center random selects some samples from the large dataset to send to each client. The in-processing training is to train the client model on its private dataset and the aggregated dataset received from the center. The aggregated dataset received by each client from the center contains generated samples from other clients. These samples and their corresponding labels can be regarded as other clients’ classification result on an unlabeled sample, because the samples are generated by other clients according to the target labels. At this point, in-processing training with the aggregated dataset is equivalent to training with the results of labeling unlabeled data with another classifier in co-training.

\subsection{Algorithm Analysis}
PerFED-GAN is a personalized federated learning scheme, which is characterized by supporting clients with different model architectures, and its detailed algorithm flow is shown in Algorithm 1. $M_i$ is the model architectures and $w_i$ is the corresponding model parameters of $M_i$. $GM_i$ is the generator network with its parameters $v_i$. $Tr_i$ is the local training dataset of Client $C_i$. MAXROUND is the maximum communication rounds. $A_i$ is the dataset containing the generated samples by $GM_i$ with their generating labels. $Ca$ is a dataset allocated in center server for storing the samples in $A_i$ uploaded by each client. $Tr_i^{'}$ is a dataset containing all samples in $Tr_i$ and the random selected samples from $Ca$.

Some parts of the PerFED-GAN algorithm are independently replaceable, including local model initialization training methods, GAN training methods and local model update training methods. In theory, any method that can be used to train a neural network can be used as a replacement. Therefore, PerFED-GAN can take advantage of the latest neural network optimization research results to enhance the training effect and hardly need to make more additional changes.

The client-side execution part of the algorithm can be performed by different clients in parallel and asynchronously, and in the task executed by the central server, the time complexity of merging the datasets uploaded by each client is mainly proportional to the total number of samples uploaded by all clients. The number of uploaded samples can be adjusted according to actual conditions, and more samples can be uploaded when the communication bandwidth is sufficient. In addition, it is also a feasible strategy to directly upload the generation network to the center and generate samples by the center, but it may also bring additional risk of model privacy leakage. In addition, the impact of uploading a generative model or generating samples on the efficiency of training time should also be evaluated according to various factors such as communication capacity, sample size, size of the generative model, and computing resources of the central server.
\end{tred}

\begin{algorithm}[!t]
\begin{tred}
\DontPrintSemicolon
 \caption{PerFED-GAN}
 \KwIn{$\{M_i(w_i),G_i(v_i),Tr_i|1<=i<=N\}$}
 \KwOut{$\{w_i|1<=i<=N\}$}
 \tcp*[l]{Clients}
 \For {$i = 1$ \textbf{to} $N$ parallel}
 {
 Update $w_i$ by local training $M_i$ on $Tr_i$.\;
 }
 \For {$round = 1$ \textbf{to} MAXROUND}
 {
 
     \tcp*[l]{Clients}
     \For {$i = 1$ \textbf{to} $N$ parallel}
     {Update $w_i, v_i$ by training the GAN consists of $M_i$(discriminator) and $GM_i$(generator) on $Tr_i$. \;
     Generate dataset $A_i$ from random seeds by $GM_i(v_i)$.\;
     Upload $A_i$ to center server.\;
     }
 
     \tcp*[l]{Center}
     $Ca$ = EMPTY SET containing all classes.\;
     \For {$i = 1$ \textbf{to} $N$}
     {
     Merge $A_i$ into $Ca$ according to the class of each samples of $A_i$.\;
     }
     \For {$i = 1$ \textbf{to} $N$}
     {
     Randomly select $n_s$ samples from $Ca$ and send them to the Client $i$.\;
     }
 
     \tcp*[l]{Clients}
     \For {$i = 1$ \textbf{to} $N$ parallel}
     {Receive and merge the $n_s$ samples from center to $Tr_i$ to get $Tr_i^{'}$.\;
     Update $w_i$ by local training $M_i$ on $Tr_i^{'}$.\;
     }
 }
 \Return{$\{w_i|1<=i<=N\}$}
 \end{tred}
\end{algorithm}

\subsection{Convergence Analysis}
Given $\mathcal{X}$ is a input space, and $\mathcal{Y}$ is a output space. $\mathcal{X}$ is with distribution $\mathcal{D}$. For simplicity, suppose $\mathcal{Y}=\{-1,+1\}$. Let $\mathcal{H}:\mathcal{X} \rightarrow \mathcal{Y}$ denote the hypothesis space, $ \mid \mathcal{H} \mid < \infty $, and the ground truth $h \in \mathcal{H}$, which means the generalization error of $h$ is 0.

\textbf{Definition 1.} Suppose we have two classifiers $h_i, h_j \in \mathcal{H}$, we define the generalization disagreement between them as:
\begin{equation}
    \begin{aligned}
     d(h_i,h_j)&=d(h_i,h_j \mid \mathcal{X}) \\
               &=\mathbf{Pr}(h_i(x) \ne h_j(x) \mid x \in \mathcal{X})
    \end{aligned}
\end{equation}
Obviously, for two classifier $h_i, h_j$,
$$d(h_i,h_j)=d(h_j,h_i)$$
$$d(h_i,h_i)=0$$
and for $h_i$, its generalization error can be expressed as $d(h_i, h)$. If 
\begin{equation}
\mathbf{Pr}(d(h_i, h)>=\epsilon)<\delta
\end{equation}
where $\epsilon > 0, \delta > 0$, we can say that the generalization error of the classifier $h_i$ is less than $\epsilon$ with the confidence parameter $\delta$. That is, the difference between the classifier $h_i$ and the oracle $h$ is bounded by $\epsilon$ with high probability (more than $1 - \delta$). For a certain $\delta$, a better classifier $h_i$ corresponds to a smaller $\epsilon$, and vice versa.

\textbf{Theorem 1.} Given two initial training \red{datasets} $ds_1$ of size $s_1$ and $ds_2$ of size $s_2$, the size of which are enough to train two classifier $h_1$ and $h_2$, and the generalization error of them is $a_0 < 0.5$ and $b_0 < 0.5$ respectively with the confidence parameter $\delta > 0$. A GAN using $h_1$ as a discriminator network generates $g$ number of samples, dataset $G$, and put them into \red{$D'_2$ which consists of $ds_2$ and $G$}. Then $h'_2$ is trained from $D'_2$ by minimizing the empirical risk. If
\begin{equation}
s_1 >= \frac{ln(\frac{\mid \mathcal{H} \mid}{\delta})}{a_0}
\end{equation}
\begin{equation}
s_2 >= \frac{ln(\frac{\mid \mathcal{H} \mid}{\delta})}{b_0}
\end{equation}
\begin{equation}\label{5}
e^M \sqrt{M!} - M >= s_2 b_0
\end{equation}
where $M = ga_0$, then
\begin{equation}\label{pr}
\mathbf{Pr}(d(h'_2,h)>=b_1)<\delta    
\end{equation}
where $b_1=max\{b_0+g\frac{a_0-d(h_1,h'_2)}{s_2},0\}$

\textbf{Proof.} \red{Let $d(h^i, D)$ to denote the expected rate of disagreement between $h^i$ and $D$, then,}
\begin{equation}
    d(h, D'_2)=\frac{gd(h_1,h)}{s_2 + g}
\end{equation}
\red{where $D'_2$ consists of $ds_2$ and $G$}, and $G$ is generated by the GAN whose discriminator is $h_1$, then 
\begin{equation}
    d(h'_2, D'_2)=\frac{s_2 d(h'_2,h) + g d(h_1, h'_2)}{s_2 + g}
\end{equation}
By minimizing the empirical risk, the training process is to generate the classifier which has the smallest observed rate of disagreement on training dataset. It means that traing  $h'_2$ is equivalent to minimizing $d(h'_2, D'_2)$. In order to train a better classifier whose generalization error is bounded by $b1$, \red{the dataset $D'_2$ should guarantee that the probability of classifier whose generalization error is no less than $b_1$ has a lower observed rate of disagreement on $D'_2$ than $h$ is small enough (less than $\delta$).}

\red{The generalization error of classifier $h_1$ is upper bounded by $a_0$ with confidence parameter $\delta$, so $d(h, D'_2)$ is no bigger than $\frac{ga_0}{s_1+g}$.} The probability that the classifier $h'_2$ has a lower observed rate of disagreement on $D'_2$ than $h$ is less than
\begin{equation}\label{eq1}
    C^M_{s_2+g}d(h'_2,D'_2)^M(1-d(h'_2,D'_2))^{s_2+g-M}
\end{equation}
Since that
\begin{equation}\label{b1}
    b_1=max\{b_0+g\frac{a_0-d(h_1,h'_2)}{s_2},0\}
\end{equation}
If $d(h'_2,D'_2)>=b_1$, then
\begin{equation}
\begin{aligned}
    d(h'_2,D'_2) &= \frac{s_2 d(h'_2,h) + g d(h_1, h'_2)}{s_2 + g} \\
                 &>= \frac{s_2 b_1 + g d(h_1, h'_2)}{s_2 + g} \\
                 &>= \frac{s_2 b_1 + g a_0}{s_2 + g} \\
                 &> \frac{M}{s_2 + g}
\end{aligned}    
\end{equation}
Since \red{Formula (\ref{eq1})} is monotonically decreasing as $d(h'_2,D'_2)$ increases when $M/(s_2+g)<d(h'_2,D'_2)<1$, and $d(h'_2,D'_2)>=b_1$, the value of \red{Formula (\ref{eq1})} is less than
\begin{equation}\label{eq2}
    C^M_{s_2+g}(\frac{s_2 b_1 + g a_0}{s_2 + g})^M(1-\frac{s_2 b_1 + g a_0}{s_2 + g})^{s_2+g-M}
\end{equation}
That is, the probability that the classifier $h'_2$ has a lower observed rate of disagreement on $D'_2$ than $h$ is less than \red{Formula (\ref{eq2})}. The value of \red{Formula (\ref{eq2})} can approximately calculated by Poisson Theorem
\begin{equation}
\begin{aligned}
    C^M_{s_2+g}(\frac{s_2 b_1 + g a_0}{s_2 + g})^M(1-\frac{s_2 b_1 + g a_0}{s_2 + g})^{s_2+g-M} \\
    \approx \frac{(s_2 b_1 + g a_0)^M}{M!}e^{-(s_2 b_1 + g a_0)}
\end{aligned}
\end{equation}
Assuming $e^M \sqrt{M!} - M >= s_2 b_0$, then
\begin{equation}
    \frac{(s_2 b_1 + g a_0)^M}{M!}e^{-(s_2 b_1 + g a_0)} <= e^{-s_2 b_0}
\end{equation}
Since $s_2 >= \frac{ln(\frac{\mid \mathcal{H} \mid}{\delta})}{b_0}$, then
\begin{equation}
    e^{-s_2 b_0} <= \frac{\delta}{\mid \mathcal{H} \mid}
\end{equation}
Considering that there are at most $\mid \mathcal{H} \mid - 1$ classifiers having generalization error no less than $b_1$ whose observed rate of disagreement with $D'_2$ is lower than $h$. Therefore
\begin{equation}
\begin{aligned}
    \mathbf{Pr}(d(h'_2,h)>=b_1) &<= (\mid \mathcal{H} \mid - 1) e^{-s_2 b_0} \\
                                &<= (\mid \mathcal{H} \mid - 1) \frac{\delta}{\mid \mathcal{H} \mid} \\
                                &< \delta
\end{aligned}
\end{equation} 
and Theorem 1 is proved.\hfill \IEEEQEDhere

\red{From the Formulas (\ref{b1}) and (\ref{pr}), we can find that when $d(h1,h'_2)$ is bigger}, the lower bound of the generalization error of $h'_2$ is lower at the same confidence level. Since $h'_2$ is trained on $D_2$ and $G$, and $G$ is generated by the GAN whose discriminator is $h_1$, the $d(h1,h'_2)$ is mainly dependent on how great divergence between $h_1$ and $h_2$, or how different are between the training datasets of $h_1$ and $h_2$. The large diversity between training datasets are very common in personalized federated learning settings, so the condition for performance improvement is usually met. Due to the symmetry, \red{if we swap $h_1$ and $h_2$}, the same conclusion applies to the improved version of $h_1$. 

In multiple rounds of training, \red{$h_1$ and $h_2$ obtained from the previous updates in each round are used as new initial classifiers to repeat the above process, which can further improve the performance of each client's personalized model. It should be mentioned that, however, during this process, the disparity between the models of different clients will gradually diminish. According to Theorem 1, it results in a reduced model performance improvement until the model no longer improves.} Continued training may cause the models of various customers to eventually converge and lose their individualized features, which is counterproductive to the goal of generating a high-performance tailored model.

PerFED-GAN typically has considerably more than two clients participating in federated learning and training. Our convergence analysis, on the other hand, should be equally applicable. This is due to the fact that, for a given client, all other clients can be viewed as a whole or an ensemble classifier. It is currently equivalent to the two in the preceding theoretical analysis. We can determine the convergence of PerFED-GAN by performing the above study for each client.

An obvious reason for the PerFED-GAN method is that when the models of different clients are highly diverse, \red{the gap in information possessed by the models will be bigger.} As a result, each client model's information obtained from others contains more unknown knowledge to itself. As a result, more fresh knowledge leads to more significant performance increases. Furthermore, when the mutual learning between several models is significant, the knowledge variety between them almost disappears. At this point, mutual learning is unlikely to give any client with new information.

\subsection{Hyper-parameter}

In Theorem 1, the left side of \red{Formula (\ref{5})} is increasing with $M=ga_0$, which means that, for a fixed $a_0$, a given $h_2$ with generalization error bounded by $b_0$ and $D_2$ of size $s_2$, a big enough size of dataset $G$ by $h_1$ is needed.

In the real application of PerFED-GAN, the size of the private training dataset of each client, that is, $s_2$ may be different. It can be seen from \red{Formula (\ref{5})} that when $s_2$ is larger, the required generated dataset needs to be larger to meet the requirements in \red{Formula (\ref{5})}. The inequality holds, so we use the hyperparameter $\beta$ to determine the size of the required generated dataset, that is
\begin{equation}
g=\beta s_2
\end{equation}
In this way, a reasonable hyperparameter $\beta$ can be set to provide clients with different private datasets with a sufficient number of generated samples, \red{while avoiding the high communication cost caused by excessive sample generation requirements.}

\red{\subsection{PerFED-GAN with Differential Privacy}}
\red{GANs may generate samples that are similar to the original learning samples, leading to the potential for data privacy leakage for generated data uploaded into PerFED-GAN. In the experimental section, we show that the quality of the generated samples can be reduced by limiting the GAN training rounds to reduce the risk of leaking the raw data, and the experiment results show that reducing the GAN training rounds results in a small degradation in model quality. In addition, a further measure to protect the privacy of raw data is the introduction of differential privacy GANs. Torkzadehmahani \textit{et al.}\cite{torkzadehmahani2019dp} proposed a differential privacy conditional GAN, DP-CGAN, which can be used to generate data of specified classes and preserve the privacy of training data. Therefore, the GAN training module in PerFED-GAN can be replaced with DP-CGAN or a GAN training algorithm with similar DP technologies, which can effectively avoid generating samples to leak training data privacy.}

\hfill
\section{Experiments}
In this section, we apply the PerFED-GAN method in a variety of federated learning settings to investigate the impact of various conditions and \red{compare it with existing federated learning methods.}
\subsection{Datasets}
\textbf{CIFAR10} and \textbf{CIFAR100} \cite{krizhevsky2009learning}: The CIFAR10 dataset consists of 32x32 RGB images in 10 classes, with 6,000 images per class. There are 5,000 training images and 1,000 test images per class. The CIFAR100 is just like the CIFAR10, except it has 100 classes containing 600 images each. There are 500 training images and 100 testing images per class. The 100 classes in the CIFAR100 are grouped into 20 superclasses. Each superclass consists of 5 classes, so there are 2,500 training images and 500 testing images per superclass. The classes in CIFAR10 are completely mutually exclusive, \red{as well as in CIFAR100}. 

{\textbf{FEMNIST}\cite{caldas2018leaf}: LEAF is a modular benchmark framework for learning in federated settings. FEMNIST is an image dataset for the classification task of LEAF. FEMNIST consists of 805,263 samples of handwritten characters (62 different classes including 10 digits, 26 uppercase and lowercase English characters) from 3550 users.

\subsection{Data Settings}
\textbf{IID Data Setting}: \red{In the IID context, it is typically assumed that the data distributions of different clients' local datasets are similar and independent, i.e., they are independent and identically distributed.} Using the CIFAR10 dataset as an example, the method for constructing a private dataset under its IID settings is to randomly sample each client with a uniform distribution from the CIFAR10 training set. To avoid the interference of overlapping samples, \red{sampling without replacement is used to ensure that no overlapping samples exist among each client's training datasets.}

\textbf{Non-IID Data Setting}: The IID environment is an overly perfect federated learning environment. Existing federated machine learning approaches typically outperform in this context. However, in practice, the data assumption is far too ideal, particularly in personalized federated learning settings. In contrast to IID configuration, the distribution of the client's local private data does not adhere to independent and identical distribution under \red{Non-IID} setting, and there may be significant discrepancies between them. As an example, considering the CIFAR10 dataset, in the case of \red{Non-IID} setting, some of the clients may have a very small number of samples of a specific category, while other clients have a large number of samples of that category. In extreme cases, there may even be pathological distribution differences, \red{e.g.}, some clients have no samples of category {\it{A}} at all, while other clients have no samples of category {\it{B}} at all. Existing federated learning methods usually perform poorly when dealing with Non-IID situations, the convergence speed is greatly reduced, and \red{the final results are poor}. Therefore, the setting of Non-IID data is generally considered to be more challenging than the setting of IID data, and this is also a problem that has to be faced in personalized federated learning.

\subsection{Model Settings}
The PerFED-GAN method enables clients to design neural networks with different architecture independently. We use convolution neural networks (CNNs) with different architectures as the personalized model for each client. The model architecture of each client is a randomly generated 2-layer or 3-layer convolutional neural network, using ReLU as the activation function with a following \red{2×2} max pooling layer. The number of filters in the convolutional layer is randomly selected from \{20, 24, 32, 40, 48, 56, 80, 96\} with an ascending order. The global average pooling layer and fully connected layer (dense layer) are insert before the softmax layer of each network.} In our experiment, \red{100 clients participate in federated learning training. Table \ref{net} shows the design parameters of 10 of the network structures as examples.}

\begin{table}[ht]
\caption{Network Architectures}
\label{net}
\centering
\begin{tabular}{cccc}
\toprule
Model & \begin{tabular}[c]{@{}c@{}}1st \\ conv layer\end{tabular} & \ \begin{tabular}[c]{@{}c@{}}2nd \\ conv layer\end{tabular} & \ \begin{tabular}[c]{@{}c@{}}3rd \\ conv layer\end{tabular} \\ \midrule
1 & 24 \red{3×3} filters & 40 \red{3×3} filters & none \\
2 & 24 \red{3×3} filters & 32 \red{3×3} filters & 56 \red{3×3} filters \\ 
3 & 20 \red{3×3} filters & 32 \red{3×3} filters & none \\ 
4 & 24 \red{3×3} filters & 40 \red{3×3} filters & 56 \red{3×3} filters \\ 
5 & 20 \red{3×3} filters & 32 \red{3×3} filters & 64 \red{3×3} filters \\ 
6 & 24 \red{3×3} filters & 32 \red{3×3} filters & 64 \red{3×3} filters \\ 
7 & 32 \red{3×3} filters & 32 \red{3×3} filters & none \\ 
8 & 40 \red{3×3} filters & 56 \red{3×3} filters & none \\ 
9 & 32 \red{3×3} filters & 48 \red{3×3} filters & none \\ 
10 & 48 \red{3×3} filters & 56 \red{3×3} filters & 96 \red{3×3} filters \\ \bottomrule
\end{tabular}
\end{table}

\subsection{GAN Settings}
In each set of experiments, \red{we equip each client with a corresponding GAN model to generate samples representing the characteristics of their data distribution.} The structures of these models are independent of each other and do not need to be shared with other clients. \red{In the GAN model of a client}, the discriminant model is the client's own demand model, and the generative model is similar in structure to the discriminant model.

\subsection{Performance Evaluation}
\red{The motivation of federated learning is to obtain a higher-quality model than the local training does}, so relative test accuracy (RTA) can be used to measure the improvement in model quality. For example, the test classification accuracy of the local model is 60\%, and the test accuracy of the model obtained through federated learning on the same test set is 80\%, then the relative test accuracy of the federated learning method for the client is 80\%/60\% $\approx$ 1.33. \red{In our experiments, the performance of a federated learning method is evaluated by mean relative test accuracy(MRTA), which is the average of the relative test accuracy of the federated learning models of all clients
\begin{equation}
    \text{MRTA}=\frac{1}{N}\underset{i=1}{\overset{N}{\sum}}{ac_i}
\end{equation}
where the $N$ is the number of clients, and $ac_i$ is the relative test accuracy of Client $i$.}

\subsection{\red{CIFAR Dataset Experiments}}
\red{In the CIFAR experiments, we divided the datasets of CIFAR10 and CIFAR100 respectively. The divisions are carried out in two ways, i.e., the above-mentioned IID setting and Non-IID setting. In the experiments, for a client, the local training dataset size is 500. Under IID settings, the training sets of CIFAR10 and CIFAR100 are randomly and evenly distributed to 100 clients. For \red{Non-IID} setting, the training dataset of a client is constructed as follows: randomly select 60\% of all categories (6 categories for CIFAR10, 60 categories for CIFAR100), and uniformly random samples from these categories 450 samples, and the remaining 50 samples are randomly selected from the remaining 40\% of the samples.} The result of this configuration is that each client has a relatively large number of training samples for 60\% of the categories, while the number of training samples for the remaining 40\% of the categories is insufficient (or even missing). It should be noted that in both IID setting and \red{Non-IID} setting, since sampling without replacement is used, all clients are ensured that there are no overlapping training samples.

\red{For the test set used for evaluation, in the case of IID, it can be considered that the task demand tendencies of all clients are the same because their training dataset distributions are similar.} So we use all the samples in the test set of CIFAR10 and CIFAR100 to test the model of each client. \red{In the case of Non-IID, the needs of each customer are personalized, which is reflected by the differences in the distributions of their training datasets. Therefore, in order to test the performance of the personalized model, its test dataset distribution should be similar to that of the training dataset.} This means that for a specific client, all test samples cannot be used directly, but some types of test samples are reduced to make the test dataset and \red{the} distribution of the training dataset is consistent.

\red{In addition, in the experiments, we set the hyperparameter $\beta$ as 5, that is, each client can obtain a generated dataset with 5 times the number of local training samples.}

\begin{figure}[ht]
\centering
\subfloat[\centering \color{black}{IID setting}]{\includegraphics[width=2.6in]{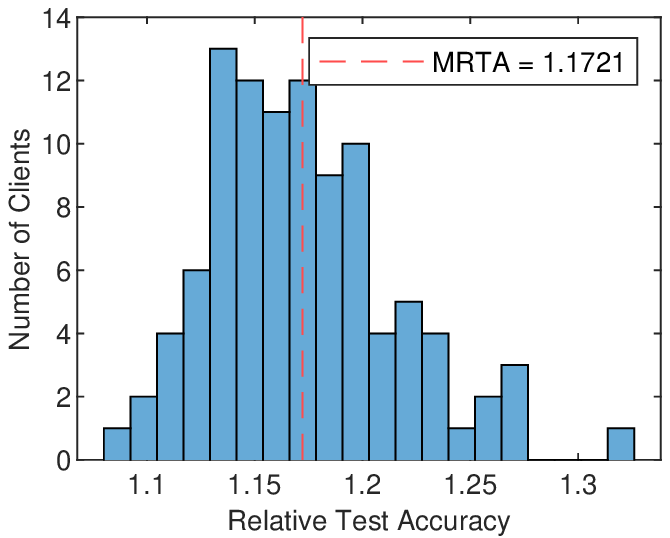}}
\qquad
\subfloat[\centering \color{black}{Non-IID setting}]{\includegraphics[width=2.6in]{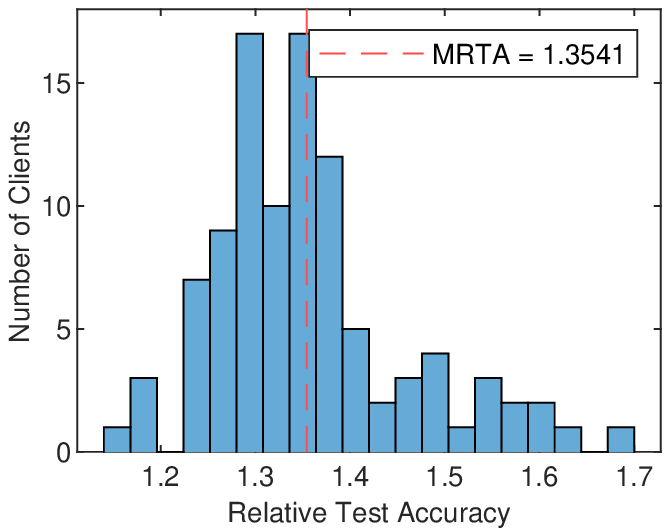}}
\caption{\red{Results of CIFAR10 experiments.}}
\label{cifar10}
\end{figure}

\begin{figure}[ht]
\centering
\subfloat[\centering \color{black}{IID setting}]{\includegraphics[width=2.6in]{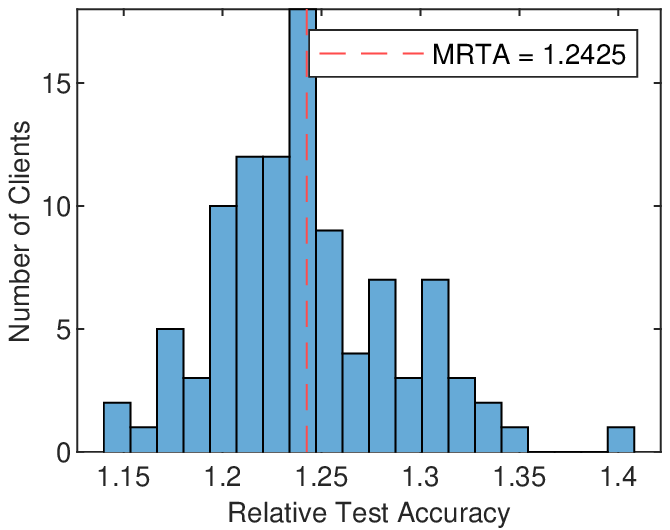}}
\qquad
\subfloat[\centering \color{black}{Non-IID setting}]{\includegraphics[width=2.6in]{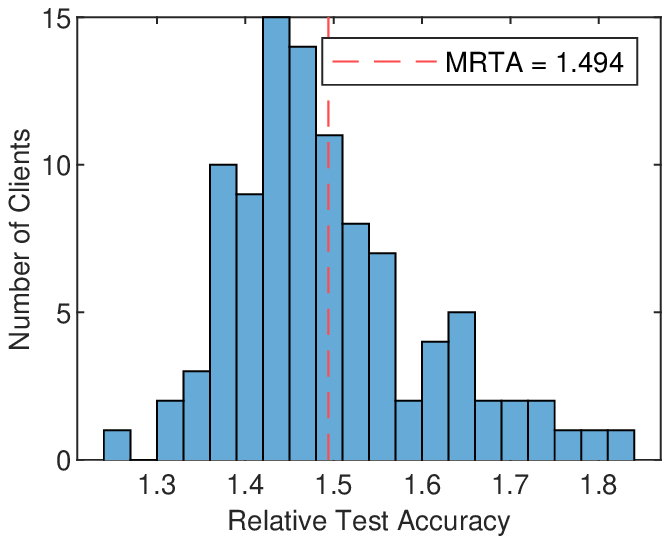}}
\caption{\red{Results of CIFAR100 experiments.}}
\label{cifar100}
\end{figure}

We respectively compared the \red{accuracies} of the local training models of CIFAR10 and CIFAR100 under IID and \red{Non-IID} settings, \red{and the results are shown in Fig. \ref{cifar10} and Fig. \ref{cifar100}. We find that the PerFED-GAN method improves the test accuracies of client models by 8\%-33\% in the IID setting with an average of 17.2\% for CIFAR10 experiments. It shows that PerFED-GAN can bring significant performance improvement of client models even when the model differences are small. For CIFAR100 experiments, the performances of PerFED-GAN model relative to the local models have been improved by 15\%-41\% under IID setting, with an average of 24\%.} This improvement is more obvious than that in the CIFAR10 experiment due to the fact that the number of samples available for training in each category is fewer, so the benefits of federated learning are greater. For the \red{Non-IID} data setting, the improvement by PerFED-GAN can be more significant due to the greater diversity among different client models since their training dataset come from different distributions. In our \red{experiments}, the PerFED-GAN achieves a relative improvement in the average test accuracy of 35\% for CIFAR10 and 49\% for CIFAR100. For each client, the improvement ranges from 14\% to 67\% for CIFAR10, and 22\%-85\% for CIFAR100. This result is significantly \red{better} than the IID settings, indicating that PerFED-GAN is less damaged by statistical heterogeneity. Therefore, PerFED-GAN is more suitable for personalized federated learning scenarios where the client tasks are more personalized and the data distribution is more different.

\subsection{FEMNIST Dataset Experiment}

Different from the CIFAR dataset which is \red{a standard benchmark for the general machine learning}, FEMNIST dataset is customized for federated learning settings. We selected 100 users with the largest number of samples from the 3550 users of FEMNIST as clients. 40\% of the samples of each client are used as training data, and the rest are test data. Each client needs to train a model to recognize the user's handwritten characters of 62 classes. 

\begin{figure}[ht]
\centering
\subfloat[\centering \color{black}{IID 
setting}]{\includegraphics[width=2.6in]{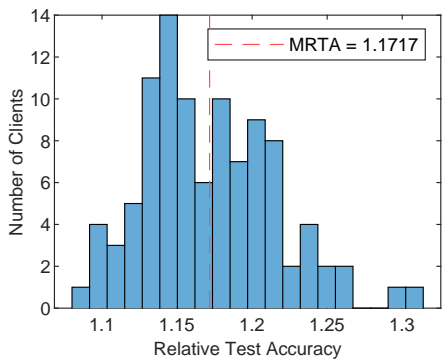}}
\qquad
\subfloat[\centering \color{black}{Non-IID setting}]{\includegraphics[width=2.6in]{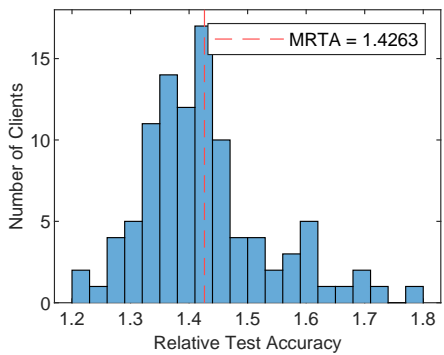}}
\caption{\red{Results of FEMNIST experiments.}}
\label{fe}
\end{figure}

\red{The training dataset setting methods in IID data settings and \red{Non-IID} settings are similar to the CIFAR experiments. The model architectures of the clients are the same as that used in the CIFAR experiments}, and the local training parameters are also tuned in a similar way as the CIFAR experiments. \red{In this experiment, we set the hyper-parameter $\beta$ as 5, and the results are shown in Fig. \ref{fe}. We find that the PerFED-GAN can improve the test accuracies of almost all clients with an average of 17\%, within range 9\% to 31\% for IID setting, and an average of 43\% within range 21\%-80\% for Non-IID setting.}

\subsection{Hyper-parameter $\beta$}

According to theoretical analysis, PerFED-GAN needs enough generated samples for each client, and the number of samples required increases with the number of local private training samples. The meaning of the hyperparameter $\beta$ is the ratio of the number of generated samples provided to the number of local training samples. \red{A too small $\beta$} value may weaken the training effect of PerFED-GAN, while a too large $\beta$ value may bring excessive communication costs. In this part, we test the effect of different $\beta$ value settings on the effect of \red{PerFED-GAN method}.

\begin{figure}[ht]
\centering
\includegraphics[width=3in]{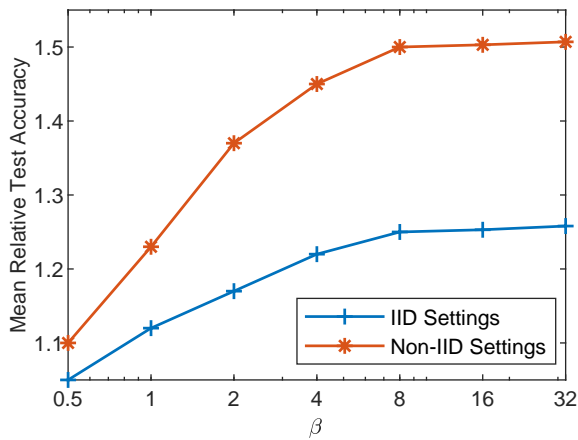}
\caption{\red{$\beta$ v.s. MRTA on CIFAR100 experiments.}}
\label{beta}
\end{figure}

We repeat the CIFAR100 \red{experiments} with different values of $\beta$ and record the changes of performance of PerFED-GAN. The results are shown in Fig.\ref{beta}. 

The results show that when the value of $\beta$ increases, PerFED-GAN can indeed provide higher performance, but its impact on the PerFED-GAN is not monotonous. This benefit decreases as $\beta$ increases. At this time, the number of generated samples required increases in proportion to the $\beta$. It is also an increase in communication overhead. For example, when the $\beta$ value is 32, it means that each client needs 32 times the generated samples of its local \red{dataset}, which greatly increases the communication cost, \red{and the impovement is only 2\% increase compared to the result when $\beta$ is 8}, but the communication overhead is increased by 300\%.

In practical applications, communication conditions and costs should be considered, and an appropriate hyperparameter $\beta$ value should be selected to make a trade-off between the model performance gain and communication cost brought by PerFED-GAN.

\subsection{Round of GAN Training}

\begin{figure}[ht]
\centering
\includegraphics[width=3in]{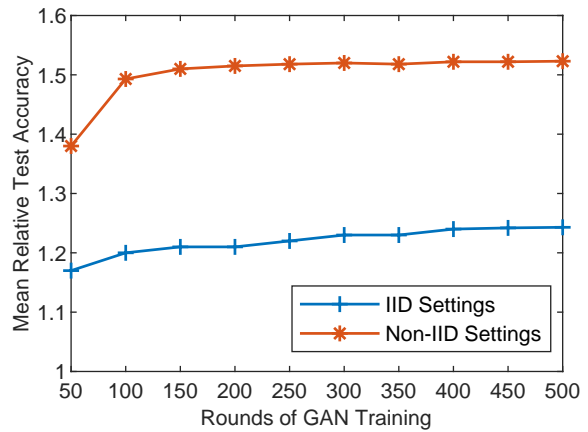}
\caption{\red{Round of GAN Training v.s. MRTA on CIFAR100 experiments.}}
\label{gan}
\end{figure}

\red{From theoretical analysis, it can be known that when the samples generated by GAN are more similar to the training samples, the more accurately the generated samples can express the information of the training samples. As a result, other client models that use these generated samples can learn the information more accurately, which brings more significant performance gains to the federated learning model. However, high-quality generated samples may also cause the leakage of training samples. Avoiding data privacy leakage is the basic premise of federated learning. Therefore,  we try to control the quality of generated samples of GANs by training rounds. When the training rounds are insufficient, the qualities of generated samples are lower, which can better protect the privacy of clients' local data. We investigate whether using these low-quality generated samples with PerFED-GAN can result in sufficient federated learning performance gains.}

\red{We repeated the experimental process on CIFAR100, with different GAN training rounds to detect its impact on the final performance of PerFED-GAN, using MRTA for the performance evaluation. The experiment results are shown in Fig. \ref{gan}. The experimental results show that higher-quality generated samples obtained after more rounds of GAN training can indeed improve PerFED-GAN performance, but even if the GANs obtained from less training rounds are used to generate samples, when the qualities of the samples generated at this time are very low and the risks of privacy leakage are low, they can still achieve better results. As a result, the effect is not much different from that of more rounds.}

Therefore, in practical applications, PerFED-GAN only needs to use lower-quality GANs to generate samples, which protects local private data and reduces the computational overhead and \red{time cost in local training of GANs.}

\red{\subsection{DP-CGAN Experiments}}
\red{In addition to directly controlling the number of GAN training rounds to reduce the risk of generating data leakage of training data privacy, we also conduct experiments in which using DP-CGAN for GAN training. According to the study in [5], DP-CGAN can effectively reduce the risk of exposing training data privacy while maintaining high generation quality.}

\begin{figure}[ht]
\centering
\subfloat[\centering \color{black}{IID setting}]{\includegraphics[width=2.6in]{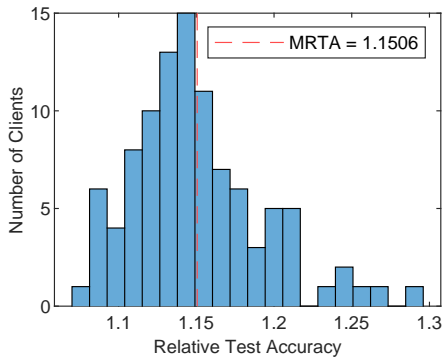}}
\qquad
\subfloat[\centering \color{black}{Non-IID setting}]{\includegraphics[width=2.6in]{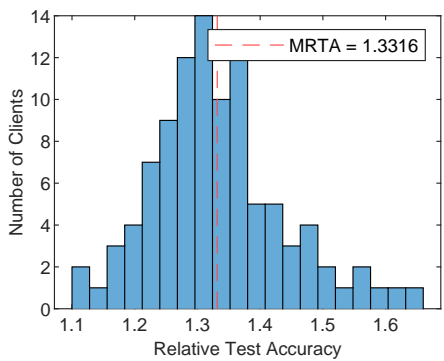}}
\caption{\red{DP-CGAN Results of CIFAR10 experiments.}}
\label{cifar10DP}
\end{figure}

\begin{figure}[ht]
\centering
\subfloat[\centering \color{black}{IID setting}]{\includegraphics[width=2.6in]{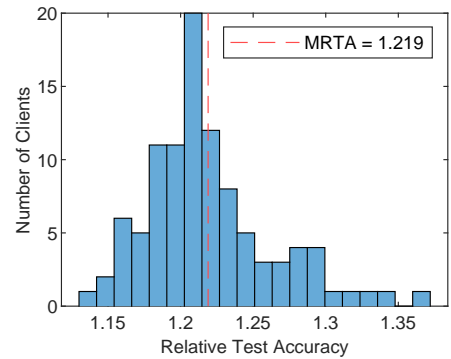}}
\qquad
\subfloat[\centering \color{black}{Non-IID setting}]{\includegraphics[width=2.6in]{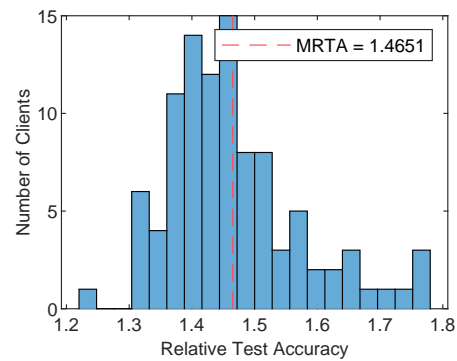}}
\caption{\red{DP-CGAN Results of CIFAR100 experiments.}}
\label{cifar100DP}
\end{figure}

\red{We repeated the experimental process of CIFAR10 and CIFAR100, but replaced the GAN training part of the original algorithm with DP-CGAN, using $\delta=10^{-5}$, to detect the impact of the replacement algorithm on the final performance of PerFED-GAN, using MRTA for performance evaluation. The experiment results in Fig. \ref{cifar10DP} and Fig. \ref{cifar100DP} show that, after the application of DP-CGAN with stronger privacy protection ability, the average accuracy of the personalized federated learning of PerFED-GAN decreases, but the magnitude is small, indicating that the proposed algorithm brings a small performance loss cost with stronger data privacy security protection.}

\hfill
\subsection{Compare with Other Federated Learning Methods}
PerFED-GAN is a personalized federated learning method that tries to be compatible with heterogeneous models which have different network architectures. Therefore, there are few federated learning methods to compare with it directly. 

We use personalized FedAvg method \cite{deng2020adaptive} for comparison, and it \red{fine-tunes} the global federated model to each client to perform personalized tasks.
At the same time, we also try to apply the above ideas to the FedProx method \cite{li2018federated}, which is considered to perform better in the case of \red{Non-IID} data distribution. For the methods compatible with different network structures, we choose the FedMD method \cite{li2019fedmd} for comparison. This method uses model distillation and a proxy dataset to align different client models on the proxy dataset to achieve personalized federated learning.

\red{We use the data configuration of CIFAR100 experiments in this subsection.} Considering that the personalized FedAvg and FedProx methods \red{cannot} support models with different architectures, we use 100 clients having neural network models with the same architecture and different parameter values \red{for comparison. In the experiment for comparing with FedMD, we adopt the same experiment setup as our CIFAR100 experiments, that is, we selected 100 clients with different architectures of neural networks.}

\red{We tried different hyperparameter settings for better performance in personalized FedAvg and FedProx experiments. In FedMD experiments}, we use the training settings suggested by \cite{li2019fedmd} since it is similar to our experimental conditions, i.e., using the training set of CIFAR10 as the public dataset and using 5000 samples in each round for model alignment. 

\begin{figure}[ht]
\centering
\subfloat[\centering \color{black}{IID setting}]{\includegraphics[width=2.3in]{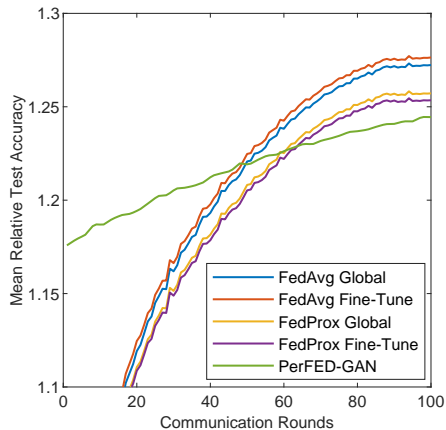}}
\qquad
\subfloat[\centering \color{black}{Non-IID setting}]{\includegraphics[width=2.3in]{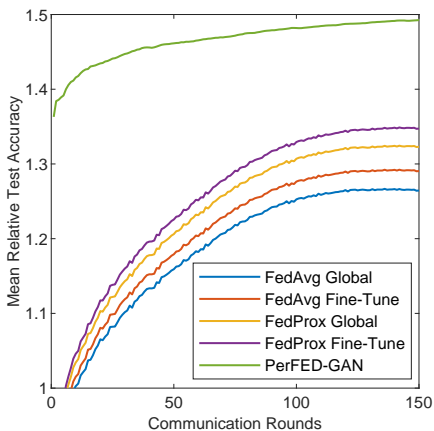}}
\caption{\red{Method comparisons for same model architecture. Algorithms FedAvg-based and FedProx-based methods are only compatible with federated learning scenarios where all clients have the same model architecture.}}
\label{fedavg}
\end{figure}

\begin{figure}[ht]
\centering
\subfloat[\centering \color{black}{IID setting}]{\includegraphics[width=2.3in]{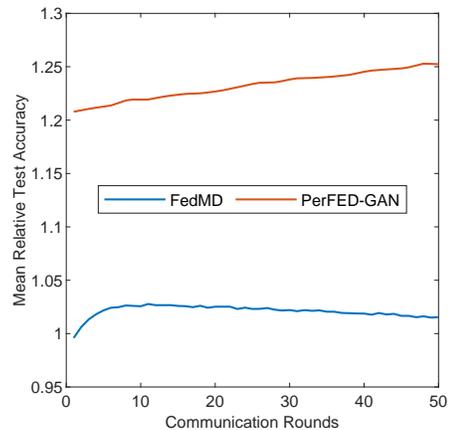}}
\qquad
\subfloat[\centering \color{black}{Non-IID setting}]{\includegraphics[width=2.3in]{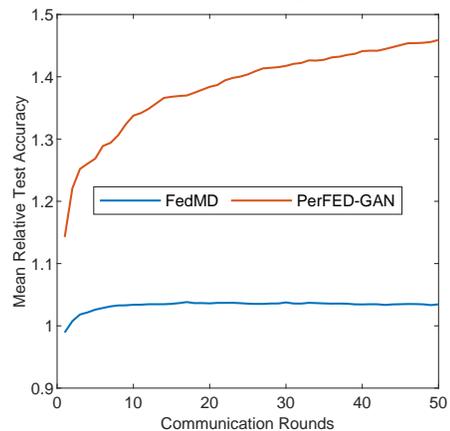}}
\caption{\red{Method comparisons for different model architectures. FedMD and PerFED-GAN are compatible with federated learning scenarios where clients have different model architectures.}}
\label{fedmd}
\end{figure}

The results of comparing FedAvg and FedProx with PerFED-GAN are shown in Fig.\ref{fedavg}. \red{In IID settings, FedAvg with fine-tuning achieves the best performance.} PerFED-GAN shows significantly better performance than the comparison methods at the beginning of the communication rounds. \red{Although PerFED-GAN fails to achieve the best performance after 100 rounds of communication, this is expected}. Because PerFED-GAN is mainly designed for personalized federated learning with large differences in data distribution. In \red{Non-IID} settings, FedAvg and FedProx both need more communication rounds to stabilize to good performance. At this time, PerFED-GAN shows great advantages over other methods in Fig.\ref{fedavg}(b), leads 12\% to the best performance of the comparison methods. Moreover, it achieves the best performance results with fewer communication rounds. \red{In terms of time consumption, the PerFED-GAN method needs to train the GAN and train on the aggregated dataset thus results in more time overhead than the parameter aggregation methods such as FedAvg and FedProx for in-client training. However, supporting personalized federated training of heterogeneous model architectures and better performance for \red{Non-IID} scenarios is the most important advantage of the PerFED-GAN algorithm, and it is also the main motivation for this paper.}

\red{The results of comparing FedMD with PerFED-GAN are shown in Fig. \ref{fedmd}. In both IID and \red{Non-IID} settings, FedMD does not reach the performance of PerFED-GAN. PerFED-GAN leads 21\% in IID settings and 42\% in \red{Non-IID} settings, and in the iterative process, FedMD even experienced performance degradation. Federated distillation algorithm FedMD requires an additional public dataset. In each round, each client needs to train on the public dataset to align the output probability vectors of different client models. Therefore, the additional time consumption mainly depends on the size of the public dataset used for federated distillation. In this set of FedMD experiments, the size of public dataset is 10 times the size of local private dataset. Compared with this, the training of the GAN part in the PerFED-GAN experiments in this paper only needs to be performed on the local dataset. The in-processing training uses 5 times the number of generated samples compared to the local samples, which is lower than the size of the public dataset in FedMD. From the experimental results, the quality of the personalized model of PerFED-GAN is significantly better than that of the federated distillation method FedMD.  Furthermore, it should be pointed out that PerFED-GAN does not need to rely on additional datasets. In the federated distillation algorithm, a huge public dataset is necessary, and the dataset needs to be highly related to the task of federated learning. The availability of this public dataset limits the scope of application of federated distillation method.}

As a result, when compared to other methods, PerFED-GAN is not only compatible with different network structure designs, but it is also suitable for personalized federated learning scenarios with large differences in client data \red{distributions. The model performances of our method outperforms comparison methods especially in \red{Non-IID} data settings, and the proposed method requires fewer rounds of communication can be completed more quickly.}

\section{Conclusion}
\red{In this paper, we propose the PerFED-GAN federated learning method, which is compatible with heterogeneous model architectures. PerFED-GAN is more suitable for personalized federated learning settings with higher heterogeneity and personalized needs than existing methods. Furthermore, PerFED-GAN protects not only the private data of all clients in federated learning settings, but also their private models and training strategies. PerFED-GAN enables clients to share multi-party knowledge to improve the performance of their local models. It produces promising results on \red{Non-IID} datasets for heterogeneous models with different architectures, which are more practical but typically difficult to handle in existing federated learning methods.}

\bibliographystyle{IEEEtran} 
\bibliography{IEEEabrv,bibs}

\end{document}